\pdfoutput=1
\documentclass[11pt, letterpaper, logo, onecolumn, copyright, numbering]{paper_improved}

\usepackage{soul}
\usepackage{xcolor}

\usepackage{booktabs}
\usepackage{enumitem}
\usepackage{array}
\usepackage{colortbl}
\usepackage{xcolor}
\usepackage{caption}
\usepackage{tabularray}
\usepackage[utf8]{inputenc} 
\usepackage[T1]{fontenc}    
\usepackage{hyperref}       
\usepackage{url}            
\usepackage{booktabs}       
\usepackage{amsfonts}       
\usepackage{nicefrac}       
\usepackage{microtype}      
\usepackage{xcolor}         
\usepackage{graphicx}

\usepackage{natbib}
\usepackage{verbatim}
\usepackage{inconsolata}
\usepackage{tabularx}
\usepackage{float}
\usepackage{multirow}
\usepackage{listings}
\usepackage{xcolor}
\usepackage{subfigure}
\usepackage{subcaption}
\usepackage{amsmath}
\usepackage{amssymb}
\usepackage{array}
\usepackage{svg}
\usepackage[T1]{fontenc}
\usepackage{wrapfig}

\definecolor{lightgray}{rgb}{0.95, 0.95, 0.95}
\definecolor{darkgray}{rgb}{0.4, 0.4, 0.4}
\definecolor{backcolour}{rgb}{0.95,0.95,0.92}
\definecolor{myblue}{rgb}{0.2, 0.4, 0.8} 
\definecolor{mygreen}{rgb}{0.2, 0.6, 0.2} 

\usepackage{amsmath,amssymb,amsfonts}

\usepackage{CJKutf8}
\usepackage{soul}
\usepackage{url}
\usepackage[utf8]{inputenc}
\usepackage{caption}
\usepackage{graphicx}
\usepackage{xcolor}
\usepackage{amsmath}

\usepackage{amsthm}
\usepackage{booktabs}
\usepackage{algorithm}
\usepackage{latexsym}
\usepackage{graphicx} 
\usepackage{amsmath}
\usepackage{subfigure}
\usepackage{microtype}
\usepackage{algorithmic}
\usepackage[switch]{lineno}
\usepackage[most]{tcolorbox}
\usepackage{alltt}

\definecolor{forestgreen}{rgb}{0.13, 0.55, 0.13}

\tcbset{
    aibox/.style={
        colback=white, 
        colframe=blue!75!black, 
        colbacktitle=blue!85!black, 
        coltitle=white, 
        fonttitle=\bfseries,
        enhanced,  
        drop shadow=black!50!white,  
        boxrule=1pt,  
        boxsep=10pt,  
        left=10pt,  
        right=10pt,  
        top=6pt,  
        bottom=6pt,  
        title code={\path[tcb fill frame] ([xshift=-10pt]frame.west) -- (frame.north west) -- (frame.north east) -- ([xshift=10pt]frame.east) -- cycle;},  
        attach boxed title to top left={xshift=10pt, yshift*=-\tcboxedtitleheight/2},
        boxed title style={boxrule=0pt, frame code={}}  
    }
}

\newtcolorbox{AIbox}[2][]{aibox, title=#2, #1}

\let\cite\citep

\title{Marco-Voice Technical Report}

\author[*,1]{Fengping Tian, Chenyang Lyu\textsuperscript{*}, Xuanfan Ni, Haoqin Sun, Qingjuan Li, Zhiqiang Qian, Haijun Li, Longyue Wang, Zhao Xu, Weihua Luo, Kaifu Zhang\\

\textbf{Alibaba International Digital Commerce} \\
\textsuperscript{*} Project Lead and Corresponding Author: lyuchenyang.lcy@alibaba-inc.com 
}

\begin{abstract}
This paper presents a multifunctional speech synthesis system that integrates voice cloning and emotion control speech synthesis within a unified framework. The goal of this work is to address longstanding challenges in achieving highly expressive, controllable, and natural speech generation that faithfully preserves speaker identity across diverse linguistic and emotional contexts. Our approach introduces an effective speaker-emotion disentanglement mechanism with in-batch contrastive learning, enabling independent manipulation of speaker identity and emotional style, as well as rotational emotional embedding integration method for smooth emotion control. To support comprehensive training and evaluation, we construct CSEMOTIONS, a high-quality emotional speech dataset containing 10 hours of Mandarin speech from ten professional speakers (five male and five female) across seven emotional categories. Extensive experiments demonstrate that our system, Marco-Voice, achieves substantial improvements in both objective and subjective metrics. Comprehensive evaluations and analysis were conducted, results show that Marco-Voice delivers competitive performance in terms of speech clarity and emotional richness, representing a substantial advance in the field of expressive neural speech synthesis. Our code and dataset are publicly available at \url{https://github.com/AIDC-AI/Marco-Voice} and \url{https://huggingface.co/datasets/AIDC-AI/CSEMOTIONS} respectively.
\end{abstract}

\begin{document}

\maketitle

\section{Introduction}
The field of text-to-speech (TTS) synthesis has witnessed remarkable progress in recent years, driven by advances in deep learning and the availability of large-scale speech datasets \cite{zeng2020aligntts,kim2021conditionalvariationalautoencoderadversarial_vits,shen2023naturalspeech}. Modern TTS systems now approach or even surpass human-level performance in terms of intelligibility and naturalness, making them indispensable in a wide range of applications, including virtual assistants, audiobook narration, accessibility tools, and entertainment \cite{li2024cm,du2024cosyvoice,du2024cosyvoice2scalablestreaming}.

Despite these achievements, truly human-like speech synthesis remains an open challenge~\cite{li2024spontaneous}. In natural communication, human speech is characterized by a rich interplay of speaker identity, prosodic style (intonation, rhythm, emphasis), and nuanced emotional expression \cite{tan2021surveyneuralspeechsynthesis_survey,zhang2023surveyaudiodiffusionmodels_tts_survey,barakat2024deep}. Replicating this diversity and flexibility in synthetic speech requires effective modeling and disentanglement of these factors~\cite{wang2024samoye,meng2025ds}. The motivation for this work stems from three persistent challenges in the field: 1) Entanglement of Emotion and Speaking Style: Many TTS models intertwine speaker-specific emotion with prosodic style~\cite{chen2024stylefusion}, making it difficult to independently control voice identity and manner of speaking. This limitation restricts the personalization and expressiveness of synthesized voices, particularly in applications that require voice cloning or style transfer. 2) Balancing Prosody and Emotion Consistency: Achieving both natural prosody and consistent, expressive emotional content is difficult~\cite{wu2019end,li2022cross}. Systems often excel at one aspect at the expense of the other, resulting in speech that sounds either monotonic or emotionally incongruent~\cite{li2024mm}. 3) Limitations of Conventional Emotion Modeling: Most existing TTS systems represent emotions using discrete categories (e.g., happy, sad, angry), which fails to capture the continuous and multidimensional nature of real-world emotional expression. Moreover, these methods often struggle to maintain high speaker similarity when synthesizing emotional speech, especially in voice cloning scenarios~\cite{li2021controllable,kansizoglou2022continuous}. 4) Limited Availability of High-Quality Emotional Speech Data: Existing emotional speech datasets often suffer from limited speaker diversity, inconsistent recording conditions, or insufficient emotional coverage, particularly for non-English languages~\cite{tits2020emotional,zhou2022emotional_esd,ma2024emobox}, which constrains the development and evaluation of emotional TTS systems.

Existing solutions typically address these challenges by deploying separate modules for each function, such as distinct encoders for speaker and emotion or post-hoc prosody adjustment~\cite{park2023using,jiang2024universal}. While modularization simplifies implementation, it often leads to weak interactions among features and degrades the overall synthesis quality~\cite{diatlova2023emospeech,zhu2023multi,choi2024mels}. For instance, the separation of speaker and emotion modules may result in unnatural blending or loss of speaker identity during expressive speech synthesis. Furthermore, the discrete treatment of emotions hinders the generation of subtle or mixed affective states, which are common in natural conversations.

To address these limitations, we built Marco-Voice, a TTS system unified emotional speech generation and voice cloning; and a emotional speech dataset named CSEMOTIONS. Our contributions in this paper is of two main parts:

\textbf{1. Marco-Voice Model:}
\begin{itemize}
    \item We develop a speaker-emotion disentanglement mechanism that separates speaker identity from emotional expression, enabling independent control over voice cloning and emotional style. We also proposed to employ in-batch contrastive learning to further disentangle speaker identity with emotional style feature.
    \item We implement a rotational emotion embedding integration method to obtain emotional embeddings based on rotational distance from neutral embeddings. Finally, we introduce a cross-attention mechanism that better integrates emotional information with linguistic content throughout the generation process.
\end{itemize}

\textbf{2. CSEMOTIONS Dataset:} 

\begin{itemize}
    \item We construct CSEMOTIONS, a high-quality emotional speech dataset containing approximately 10 hours of Mandarin speech from ten professional native speakers (five male, five female), all with extensive voice acting experience. The dataset covers seven distinct emotional categories. All recordings were made in professional studios to ensure high audio quality and consistent emotional expression.
    \item We also develop 100 evaluation prompts for each emotion class across both existing datasets and CSEMOTIONS in English and Chinese, enabling thorough and standard assessment of emotional synthesis performance across all supported emotion categories.
\end{itemize} 

Marco-Voice combines these innovations to deliver expressive, natural, and highly controllable speech synthesis. By integrating speaker identity, emotional style, and linguistic content within a single framework, our system achieves superior speech quality and emotional richness while expanding the potential applications of TTS technology in multilingual and interactive environments.

\section{Methodology}

Our Marco-Voice system follows a clear pipeline: input text and reference speech are processed through separate encoders, while speaker and emotion information are embedded as conditioning signals. These features are fed into a language model that generates token representations. The emotion embeddings then interact with the LM outputs through cross-attention before being passed to a flow matching module, which generates high-quality expressive speech. The system incorporates several key innovations including speaker-emotion disentanglement, contrastive emotion learning, and adaptive cross-attention mechanisms.

\subsection{System Architecture}
The overall architecture of Marco-Voice consists of four main components: (1) input encoders that process text and speech separately, (2) embedding modules that encode speaker identity and emotional style, (3) a text-to-token language model that integrates linguistic and conditioning information, and (4) a flow matching module that generates acoustic parameters for final speech synthesis. The system is designed to handle multiple types of conditioning information while maintaining independent control over different aspects of speech generation.

\begin{figure}\centering
    \includegraphics[width=0.9\linewidth]{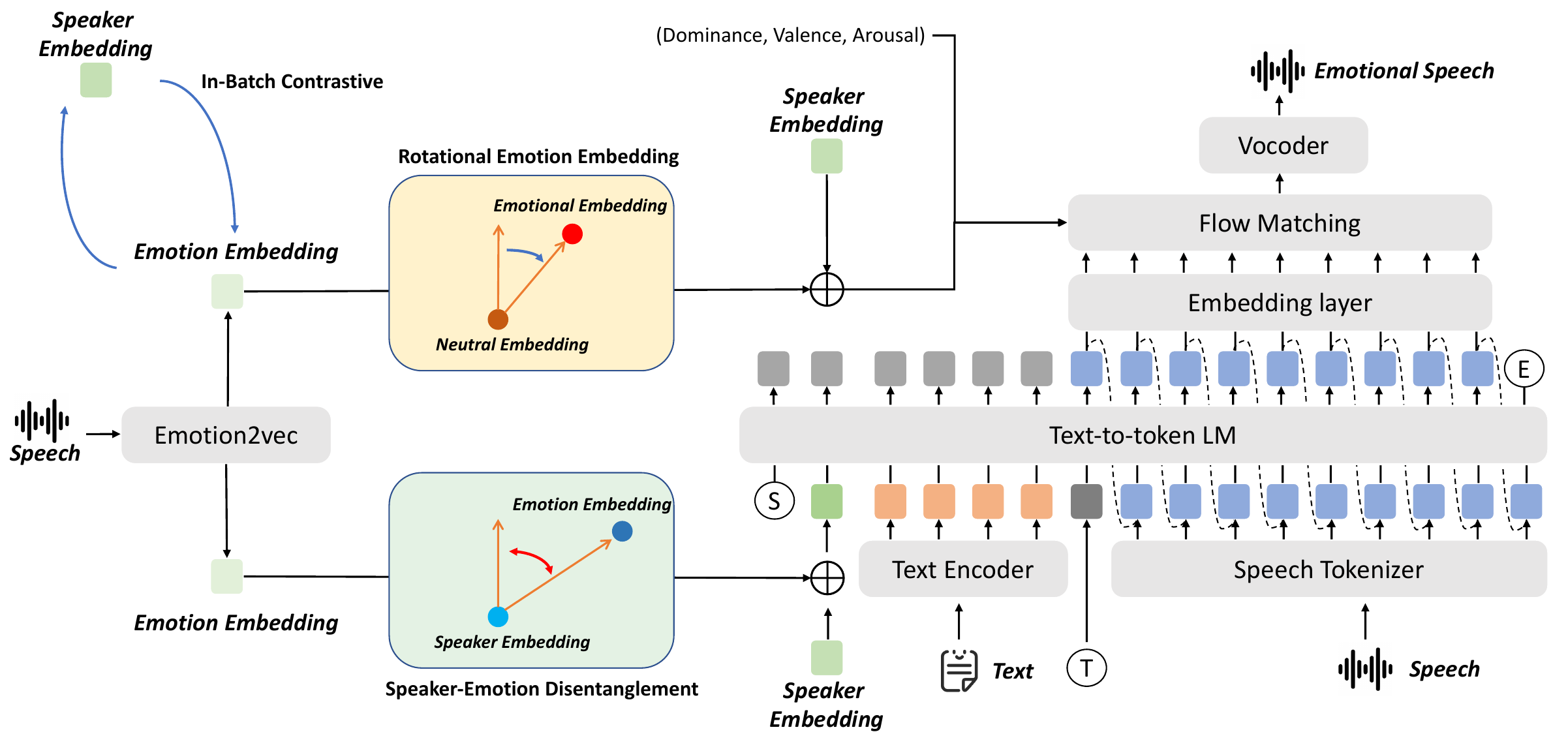}
    \caption{The overall architecture of our Marco-Voice system, incorporating speaker-emotion disentanglement, in-batch contrastive learning.}
    \label{fig:marco_voice_system}
\end{figure}

\subsection{Rotational Emotion Embedding Integration}
We use an emotion feature extractor module to extract emotion embeddings from speech. We disentangle speaker-specific qualities and speaker-independent emotion representations by using paired samples of emotional speech $x_i^e$ and neutral speech $x_i^n$ from the same speaker. These are encoded using a pre-trained emotional encoder $E_e$ to obtain representations $u_i^e = E_e(x_i^e)$ and $u_i^n = E_e(x_i^n)$. 

We adopted the method intorduced by \cite{chen-etal-2024-emoknob} that hypothesizes that the difference between these encodings captures a direction vector in the speaker embedding space corresponding to the emotional content, while removing speaker identity:
\begin{equation}
    v_i^e = \frac{u_i^e - u_i^n}{\|u_i^e - u_i^n\|}
\end{equation}
We then aggregate over $N$ such pairs to obtain a robust emotion embedding:
\begin{equation}
\label{eq:rotational_emotion_embeds}
    \mathbf{e} = \frac{1}{N} \sum_{i=1}^N v_i^e
\end{equation}
For most cases, we find that single-shot ($N=10$) suffices to produce high-quality emotion control. The resulting emotion embedding $\mathbf{e}$ serves as a conditioning signal at multiple stages, allowing the system to maintain emotional consistency from text processing through final speech generation.

\subsection{Speaker-Emotion Disentanglement}
We introduce a cross-orthogonal constraint to separate speaker identity from emotional expression. Given input features, we obtained speaker embeddings $\mathbf{s}$ using encoders $E_s$ and $E_e$:

\begin{equation}
    \mathbf{s} = E_s(\mathbf{x})
\end{equation}

where $x$ is the speech and the emotion embedding $\mathbf{e}$ is obtained in Equation. \ref{eq:rotational_emotion_embeds}. Our implementation computes the cross-orthogonality loss as follows. Given batch-wise speaker and emotion embeddings, we calculate the dot-product matrix, normalize by their vector norms, and compute the squared Frobenius norm. In addition, we calculate the average cosine similarity across all pairs in the batch, also using the squared Frobenius norm. The total orthogonality loss is a weighted sum of both terms:

\begin{equation}
\begin{aligned}
    &\text{Let}\quad S \in \mathbb{R}^{B \times D},\quad E \in \mathbb{R}^{B \times D} \\
    &\text{Dot-Product Matrix:} \quad D = E S^T \\
    &\text{Norms:} \quad n_E = \|E\|,\quad n_S = \|S\| \\
    &\text{Normalized Matrix:} \quad \hat{D} = D / (n_E n_S^T) \\
    &\mathcal{L}_{ort} = \|\hat{D}\|_F^2 + \|\text{mean}(\text{cos\_sim}(E, S))\|_F^2
\end{aligned}
\end{equation}

where $B$ is batch size, $S$ and $E$ consists of a batch of $\mathbf{s}$ and $\mathbf{e}$, $\|\cdot\|_F$ denotes the Frobenius norm, and $\text{cos\_sim}(E,S)$ computes the cosine similarity between each emotion embedding and each speaker embedding. During training, if batchwise pairwise computation is enabled, we average the absolute dot product between each embedding and all opposing emotion embeddings in the batch, except self-pairs; otherwise, we use the orthogonality loss as above. This constraint forces speaker and emotion embeddings to be perpendicular in the feature space, enabling independent control over voice identity and emotional expression.

\subsection{In-Batch Contrastive Learning}
To improve the quality of emotion representations, we employ in-batch contrastive learning \cite{gao2021simcse}. For each emotion embedding in a training batch, we encourage it to be dissimilar from other emotion embeddings that represent different emotional states.

Concretely, during training, for each pair in the minibatch, the speaker and emotion embeddings are projected and added, then, for all pairs $(i,j)$ in the batch ($i\neq j$), we accumulate the absolute dot products:
\begin{equation}
    \mathcal{L}_{contrast} = \frac{1}{N(N-1)/2} \sum_{i<j} |\langle \mathbf{h}_i, \mathbf{e}_j \rangle|
\end{equation}
where $\mathbf{h}_i$ is the sum of projected speaker and emotion embeddings for sample $i$, and $\mathbf{e}_j$ is the corresponding projected emotion embedding in the batch.

This batchwise contrastive learning encourages distinctiveness among emotion embeddings within the batch and enhances the separation of emotion representations.

\subsection{Conditional Flow Matching Module}
The conditional flow matching module \cite{lipman2023flow,tong2024improvinggeneralizingflowbasedgenerative,du2024cosyvoice} processes transforms noise into speech parameters through a continuous flow, conditioned on all the input features. Specifically, we used an additional cross-attention mechanism such that the emotion embedding serves as the query (Q), and the acoustic token outputs from the language model serve as the keys (K) and values (V):
\begin{equation}
\begin{gathered}
    Q = W_q(\mathbf{e}) \\
    K = W_k(\mathbf{h}_{LM}) \\
    V = W_v(\mathbf{h}_{LM}) \\
\end{gathered}
\end{equation}
where $\mathbf{h}_{LM}$ is the linguistic/acoustic token sequence generated by LLM, and $W_q,W_k,W_v$ are learned projections. Then we compute:
\begin{equation}
    h_{attn} = \text{Attention}(Q,K,V) = \text{softmax}\left(\frac{QK^T}{\sqrt{d_k}}\right)V
\end{equation}
with optional masking for padding positions. The output dimensions and residual connections align with the input token sequence, allowing the emotion query to dynamically modulate the linguistic representations, thus enabling emotionally coherent speech synthesis.

The flow matching module takes the attended linguistic features $\mathbf{h}_{attn}$, along with speaker and emotion conditions, to generate acoustic parameters. The module uses a combination of Transformer and ResNet1D blocks to handle both sequential dependencies and local acoustic refinements:

\begin{equation}
    \mathbf{h}_t = \text{FlowMatch}(\mathbf{h}_{attn}, \mathbf{h}_{LM}, \mathbf{e})
\end{equation}

 This approach provides stable training while allowing flexible control over the generated speech characteristics.

\subsection{Training Objective}
The overall training objective combines the main TTS loss with regularization terms for disentanglement and contrastive learning:
\begin{equation}
    \mathcal{L} = \mathcal{L}_{TTS} + \lambda_{orth} \mathcal{L}_{orth} + \lambda_{contrast} \mathcal{L}_{contrast}
\end{equation}

where $\mathcal{L}_{TTS}$ is the main speech synthesis loss, and the $\lambda$ terms control the relative importance of each constraint. The main TTS loss includes reconstruction, spectral, adversarial, and duration components to ensure high-quality speech generation.

\section{Experimental Setup}

\subsection{Datasets}

We combine established public corpora with our newly developed proprietary dataset to ensure diversity and robustness in our modeling.

\subsubsection*{Training Datasets}

\begin{itemize}
    \item \textbf{ESD (Emotional Speech Dataset):} ESD \cite{zhou2022emotional_esd} is a large-scale, high-quality resource specifically designed for emotional voice conversion and synthesis. It comprises approximately 29 hours of audio spanning five emotion categories (neutral, happy, angry, sad, and surprise) and features 20 professional speakers (10 native English and 10 native Chinese). Each speaker recorded 350 parallel utterances.
    \vspace{2pt}
    \item \textbf{CSEMOTIONS (Chinese Speech Emotions):} To supplement existing public resources, we constructed \textbf{CSEMOTIONS}, a proprietary emotional speech dataset. This corpus includes about 10 hours of Mandarin speech from ten professional native speakers (five male, five female), all with extensive voice acting experience. CSEMOTIONS covers seven distinct emotional categories, with each speaker reading a curated set of 100 Chinese and English prompts. All recordings were made in professional studios to ensure high fidelity and expressive consistency.
\end{itemize}

\subsubsection*{Evaluation Datasets}

\begin{itemize}
    \item \textbf{LibriTTS:} For evaluation on English TTS and conversion tasks, we utilize LibriTTS~\cite{zen2019libritts}, a large-scale, multi-speaker speech corpus derived from public domain audiobooks. We sample \textbf{400 prompts} from LibriTTS.
    \vspace{2pt}
    \item \textbf{AISHELL-3:} To assess Mandarin performance, we employ AISHELL-3~\cite{shi2021aishell3}, a multi-speaker Mandarin TTS dataset with approximately 85 hours of neutral speech from 218 native speakers. We select \textbf{400 prompts} from AISHELL-3 for evaluation.
    \vspace{2pt}
    \item \textbf{CSEMOTIONS:} To comprehensively evaluate emotional expressiveness, we construct \textbf{100 prompts for each emotional class} (across both ESD and CSEMOTIONS) in both English and Chinese as dedicated evaluation data. This allows for targeted assessment of emotional synthesis and conversion across all supported emotion categories.
\end{itemize}

\textbf{ESD} and \textbf{CSEMOTIONS} serve as our primary training resources, providing rich emotional diversity. \textbf{LibriTTS} and \textbf{AISHELL-3} are employed for evaluation, with 400 prompts each used to benchmark model performance in English and Mandarin, respectively. Additionally, emotion-specific evaluation is conducted using the eval set of \textbf{CSEMOTIONS} - 100 prompts per emotional class, ensuring robust and fine-grained assessment of emotional speech capabilities. All audio was preprocessed to a consistent format (24/48kHz sampling rate, 16-bit depth) and normalized to control for volume variations.

\subsection{Implementation Details}
The model was implemented based on CosyVoice1 \cite{du2024cosyvoice}  and trained on 8 NVIDIA A100 GPUs for approximately couple hours. We used the Adam optimizer with a learning rate of $1 \times 10^{-5}$ for llm part and $1 \times 10^{-4}$ for flow matching part and a cosine decay schedule. The batch size was set to 32 per GPU. For the weighting factors in the loss function, we used $\lambda_{orth} = 0.1$ and $\lambda_{rot} = 0.5$. These values were determined through a hyperparameter search on a validation set.

\subsection{Evaluation Metrics}
We evaluated our system mainly based on human evaluation with additional automatic metrics for analysis to address the challenges for evaluating emotional speech generation with voice cloning:
\begin{itemize}
    \item Speaker similarity was measured using a pre-trained speaker model \cite{speechbrain_v1,chen2024eres2netv2boostingshortdurationspeaker} that computes cosine similarity between speaker embeddings.
    \item Emotional expressiveness was evaluated through human ratings on a 5-point Likert scale.
    \item Overall speech quality was assessed using mean opinion scores (MOS) from human listeners, as well as objective metrics including Whisper-WER \cite{radford2022robust_whisper} and DNS-MOS.
\end{itemize}

\section{Results and Analysis}

To comprehensively assess the effectiveness of the proposed Marco-Voice system, which integrates both voice cloning and emotional speech generation, we conduct evaluations targeting these two core capabilities. Given the subjective nature of speech quality, speaker similarity, and emotional expressiveness, we primarily rely on human evaluation, supplemented by automatic metrics where appropriate. This approach ensures a robust and representative assessment of system performance in both naturalness and controllability.

Human evaluations were conducted using a panel of native speakers who rated different systems across several dimensions on a five-point Likert scale (higher is better). For speaker similarity, listeners compared generated speech to reference samples from target speakers. For emotional expressiveness, raters assessed the naturalness, clarity, and emotional content of synthesized utterances. In addition, direct A/B preference tests were performed, where raters listened to paired samples and indicated their preference for each pair. Each system was evaluated using the same set of prompts to ensure fairness and comparability.

\subsection{Voice Cloning Evaluation}

Table~\ref{tab:voice_cloning} summarizes the results for voice cloning capabilities, including speech clarity, rhythm and speaking speed, naturalness, overall satisfaction, and speaker similarity. Only systems supporting voice cloning are included.

\begin{table}[h]
\centering
\resizebox{0.93\linewidth}{!}{
\begin{tabular}{lccccc}
\hline
 & Speech Clarity & Rhythm \& Speed & Naturalness & Overall Satisfaction & Speaker Similarity \\
\hline
CosyVoice1 & 3.000 & 3.175 & 3.225 & 2.825 & 0.700 \\
CosyVoice2 & 3.770 & 4.090 & 3.150 & 3.330 & 0.605 \\
Marco-Voice & \textbf{4.545} & \textbf{4.290} & \textbf{4.205} & \textbf{4.430} & \textbf{0.8275} \\
\hline
\end{tabular}}
\caption{Human evaluation results for voice cloning systems. Higher scores indicate better performance.}
\label{tab:voice_cloning}
\end{table}

As shown in Table~\ref{tab:voice_cloning}, Marco-Voice consistently outperforms existing voice cloning systems across all evaluated dimensions. Notably, our system achieves the highest speaker similarity score (0.8275), demonstrating its effectiveness in preserving speaker identity. Improvements in speech clarity, rhythm, and overall satisfaction further highlight the advantages of our speaker-style disentanglement approach.

\subsection{Emotional Speech Generation Evaluation}

Table~\ref{tab:emotion_eval} presents the evaluation results for systems supporting emotional speech generation, including speech clarity, emotional expression, rhythm and speaking speed, naturalness, and overall satisfaction.

\begin{table}[h]
\centering
\resizebox{\linewidth}{!}{
\begin{tabular}{lccccc}
\hline
 & Speech Clarity & Emotional Expression & Rhythm \& Speed & Naturalness & Overall Satisfaction \\
\hline
CosyVoice2 & 3.770 & 3.240 & 4.090 & 3.150 & 3.330 \\
Marco-Voice & \textbf{4.545} & \textbf{4.225} & \textbf{4.290} & \textbf{4.205} & \textbf{4.430} \\
\hline
\end{tabular}}
\caption{Human evaluation results for emotional speech generation. Higher scores indicate better performance.}
\label{tab:emotion_eval}
\end{table}

According to Table~\ref{tab:emotion_eval}, Marco-Voice achieves the best performance in all evaluated aspects, especially in emotional expression (4.225) and overall satisfaction (4.430). These results validate the effectiveness of our emotion modeling strategy, enabling more natural and expressive emotional speech synthesis compared to CosyVoice2 \cite{du2024cosyvoice2scalablestreaming}.

\subsection{Direct Comparison (A/B) Tests}

In addition to rating-based evaluations, we conducted A/B preference tests in which listeners compared pairs of samples from Marco-Voice and competing systems using the same prompts. The results, presented in Table~\ref{tab:ab_test}, show the percentage of times Marco-Voice was preferred.

\begin{table}[h]
\centering
\begin{tabular}{lc}
\toprule
Compared Model & Marco-Voice Win Rate \\
\midrule
CosyVoice1 & 60\% (12/20) \\
CosyVoice2 & 65\% (13/20) \\

\bottomrule
\end{tabular}
\caption{A/B preference test results: percentage of times Marco-Voice was preferred in blind listening tests.}
\label{tab:ab_test}
\end{table}

Marco-Voice is consistently preferred over all baseline systems in direct listening comparisons, indicating that listeners value the emotional expressiveness and speaker similarity of our system in direct comparisons.

\subsection{Analysis Studies}

To further investigate the performance of the Marco-Voice system, we conducted detailed analysis studies using objective metrics on both English (LibriTTS) and Mandarin (AISHELL) datasets. We compared multiple versions of Marco-Voice including: \textbf{v1} incorporates rotational emotion embeddings as conditioning signals in both the LLM and flow matching module. \textbf{v2} adds the cross-orthogonal constraint to enforce speaker-emotion disentanglement, enabling independent control over voice identity and emotional expression. \textbf{v3} employs in-batch contrastive learning between emotion and speaker embeddings. \textbf{v4} uses a cross-attention mechanism between emotion embeddings and language model tokens to ensure coherent emotion-text integration.  The evaluation metrics are Word Error Rate (WER), Speaker Similarity (SS, using both SpeechBrain and ERes2Net), deletion and insertion errors (Del \& Ins), substitution errors (Sub), and DNS-MOS scores for perceptual quality.

\textbf{LibriTTS Results:} Table~\ref{tab:libri-tts} presents the results for the LibriTTS dataset. Across all Marco-Voice versions, the WER remains low and comparable to the best-performing baseline (CosyVoice1), with Marco-Voice-v4 achieving the lowest WER (11.4). Speaker similarity scores (both SS-speech brain and SS-ERes2Net) for Marco-Voice variants are consistently higher than CosyVoice2 and on par with or slightly exceeding CosyVoice1. DNS-MOS scores for Marco-Voice models are also competitive, indicating strong perceptual quality.

\begin{table}[ht]
\centering
\resizebox{\textwidth}{!}{%
\begin{tabular}{lcccccc}
\toprule
\textbf{System} & \textbf{CosyVoice1} & \textbf{CosyVoice1*} & \textbf{Marco-Voice-v1} & \textbf{Marco-Voice-v2} & \textbf{Marco-Voice-v3} & \textbf{Marco-Voice-v4} \\
\midrule
WER $\downarrow$ & 12.1 & 58.4 & 12.4 & 12.5 & 12.0 & \textbf{11.4} \\
SS (SpeechBrain) $\uparrow$ & 64.1 & 61.3 & 64.2 & \textbf{64.7} & 64.5 & 63.2 \\
SS (ERes2Net) $\uparrow$ & 80.1 & 64.2 & \textbf{80.3} & 79.5 & 80.1 & 74.3 \\
Del \& Ins $\downarrow$ & \textbf{413} & 2437 & 387.0 & 398.0 & 415.0 & 395.0 \\
Sub $\downarrow$ & 251 & 2040.0 & 251.0 & 286.0 & 251.0 & \textbf{242.0} \\
DNS-MOS $\uparrow$ & 3.899 & 3.879 & \textbf{3.926} & 3.900 & 3.923 & 3.860 \\
\bottomrule
\end{tabular}%
}
\caption{Objective evaluation of speech recognition and synthesis quality on the LibriTTS dataset. Metrics include word error rate (WER), speaker similarity (SS) using SpeechBrain and ERes2Net, error counts (Del \& Ins, Sub), and DNS-MOS for perceptual quality. Lower WER and error counts, and higher SS and DNS-MOS indicate better performance.}
\label{tab:libri-tts}
\end{table}

\begin{table}[ht]
\centering
\resizebox{\textwidth}{!}{%
\begin{tabular}{lcccccc}
\toprule
\textbf{System} & \textbf{CosyVoice1} & \textbf{CosyVoice1*} & \textbf{Marco-Voice-v1} & \textbf{Marco-Voice-v2} & \textbf{Marco-Voice-v3} & \textbf{Marco-Voice-v4} \\
\midrule
WER $\downarrow$ & \textbf{3.0} & 23.3 & 17.6 & 15.9 & 18.2 & 17.6 \\
SS (SpeechBrain) $\uparrow$ & 10.7 & 10.6 & \textbf{11.0} & 10.9 & 10.5 & 10.4 \\
SS (ERes2Net) $\uparrow$ & 73.5 & 54.5 & \textbf{73.8} & 73.2 & 73.7 & 67.6 \\
Del \& Ins $\downarrow$ & \textbf{11.0} & 170.0 & 212.0 & 211.0 & 212.0 & 218.0 \\
Sub $\downarrow$ & \textbf{97.0} & 674.0 & 485.0 & 408.0 & 496.0 & 471.0 \\
DNS-MOS $\uparrow$ & 3.673 & \textbf{3.761} & 3.687 & 3.701 & 3.689 & 3.656 \\
\bottomrule
\end{tabular}%
}
\caption{Objective evaluation of speech recognition and synthesis quality on the AISHELL dataset. Metrics are as in Table~\ref{tab:libri-tts}. Results demonstrate the effectiveness of Marco-Voice models for Mandarin emotional TTS, particularly in speaker similarity and perceptual quality. CosyVoice1* indicates that we continue training the base model on the same dataset, which typically leads to degraded WER performance and explains the higher WER observed in the Marco-Voice models.}
\label{tab:aishell}
\end{table}
\textbf{AISHELL Results:} Table~\ref{tab:aishell} shows the results on AISHELL. Here, Marco-Voice variants generally outperform CosyVoice2 in WER, though CosyVoice1 achieves the lowest WER (3.0). Speaker similarity (SS) and DNS-MOS values for Marco-Voice remain strong, with SS-ERes2Net showing clear superiority over CosyVoice2. Notably, deletion and insertion errors are higher for Marco-Voice models, which can be attributed to challenges in emotional prompt synthesis and the presence of vocalized pauses (e.g., ``ah,'' ``um'') that are often included in expressive and emotional speech but are not always reflected in text transcripts.

The observed WER values, while generally low, may appear elevated in some cases due to the inclusion of vocalized fillers and interjections inherent in natural, emotional speech. These elements are frequent in emotional prompts and can increase WER even when the generated speech is perceptually natural and expressive. Additionally, the complexity and variability of emotional text prompts pose extra challenges for TTS systems, potentially leading to more substitution, deletion, and insertion errors. Despite these difficulties, Marco-Voice demonstrates strong speaker similarity and perceptual quality across both English and Mandarin, validating the robustness and generalization of our approach in multilingual and emotionally-rich scenarios. Overall, these objective analysis studies complement our human evaluation findings, further confirming the effectiveness of Marco-Voice in both standard and emotionally challenging TTS scenarios.
\begin{figure}
    \centering
    \includegraphics[width=0.8\linewidth]{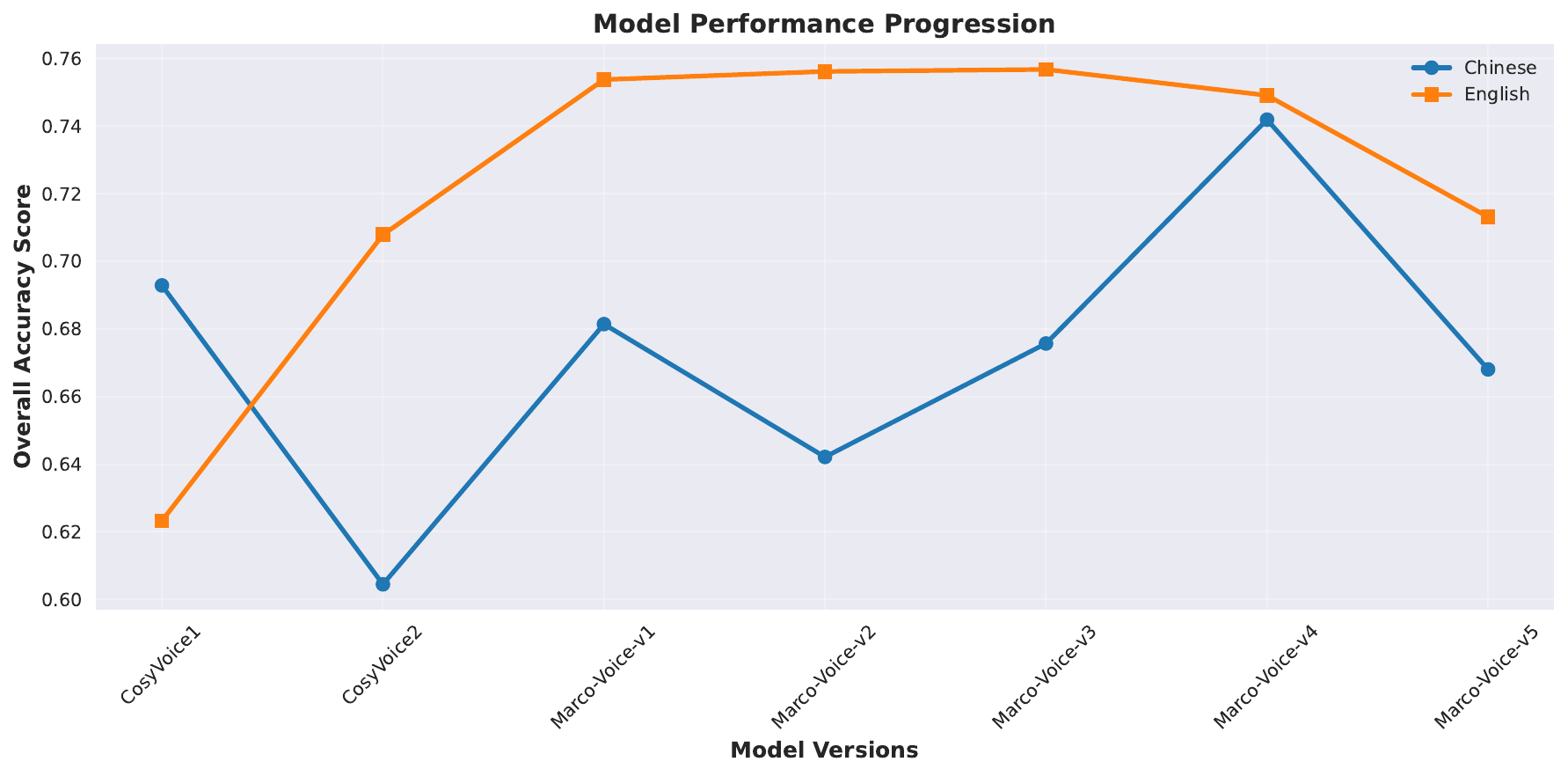}
    \caption{Overall performance progression across model versions on Chinese and English datasets. The graph shows average accuracy scores across all emotions (excluding Playfulness) for emotion recognition tasks.}
    \label{fig:overall_results}
\end{figure}

\paragraph{Model Performance Progression}
Figure~\ref{fig:overall_results} shows that Marco-Voice-v4 achieves the best performance with 0.78 accuracy on Chinese and 0.77 on English datasets. CosyVoice1 provides a strong baseline (0.72 Chinese, 0.67 English), while CosyVoice2 shows performance degradation. The Marco-Voice series demonstrates clear progression from v1 to v4, with v5 showing slight decline, indicating that v4 represents the optimal balance of architectural improvements.

\begin{figure}[t]
    \centering
    \includegraphics[width=0.96\linewidth]{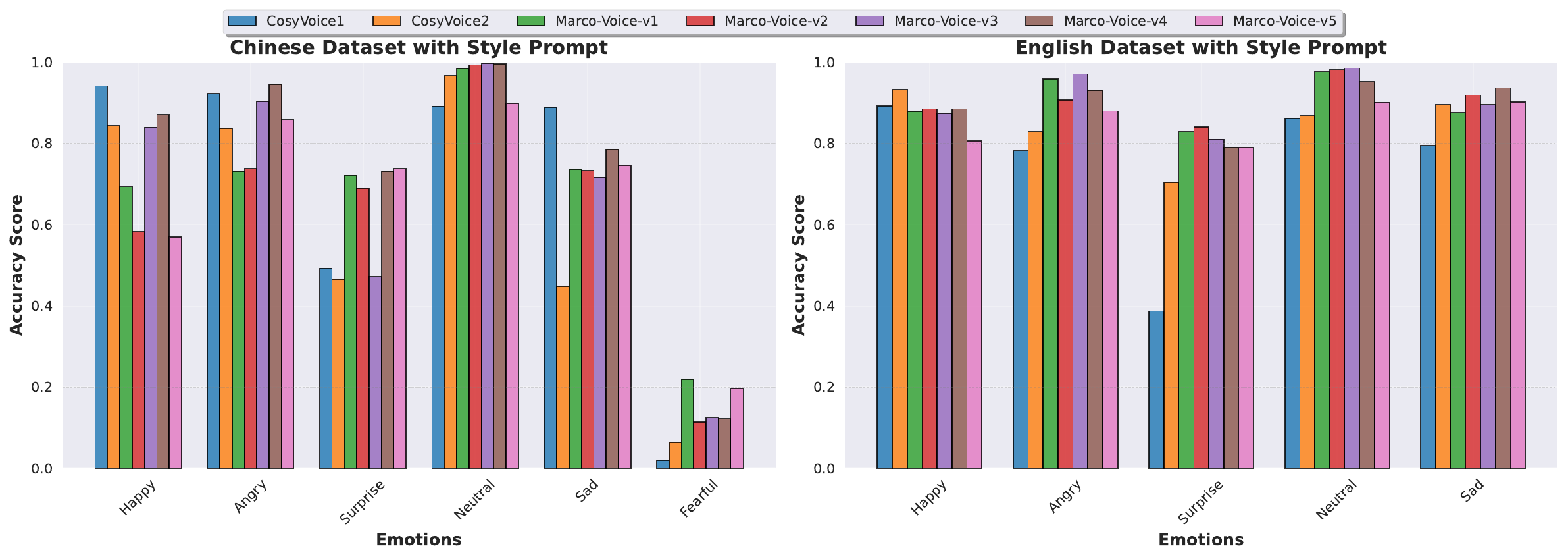}
    \caption{Emotion recognition performance comparison between Chinese and English datasets with style prompts. Results show accuracy scores for six emotion categories across seven model variants using the emotion2vec\_base\_finetuned classifier.}
    \label{fig:zh_en_results}
\end{figure}

\paragraph{Crosslingual Emotion Recognition}
Figure~\ref{fig:zh_en_results} reveals that neutral and angry emotions achieve consistently high performance (>0.85) across both languages, while surprise and sad emotions remain challenging. Marco-Voice-v4 and v5 show superior performance for complex emotions, with accuracy scores above 0.73 for surprise recognition. The relatively balanced performance between Chinese and English suggests effective crosslingual generalization.

\begin{figure}[t]
    \centering
    \includegraphics[width=0.6\linewidth]{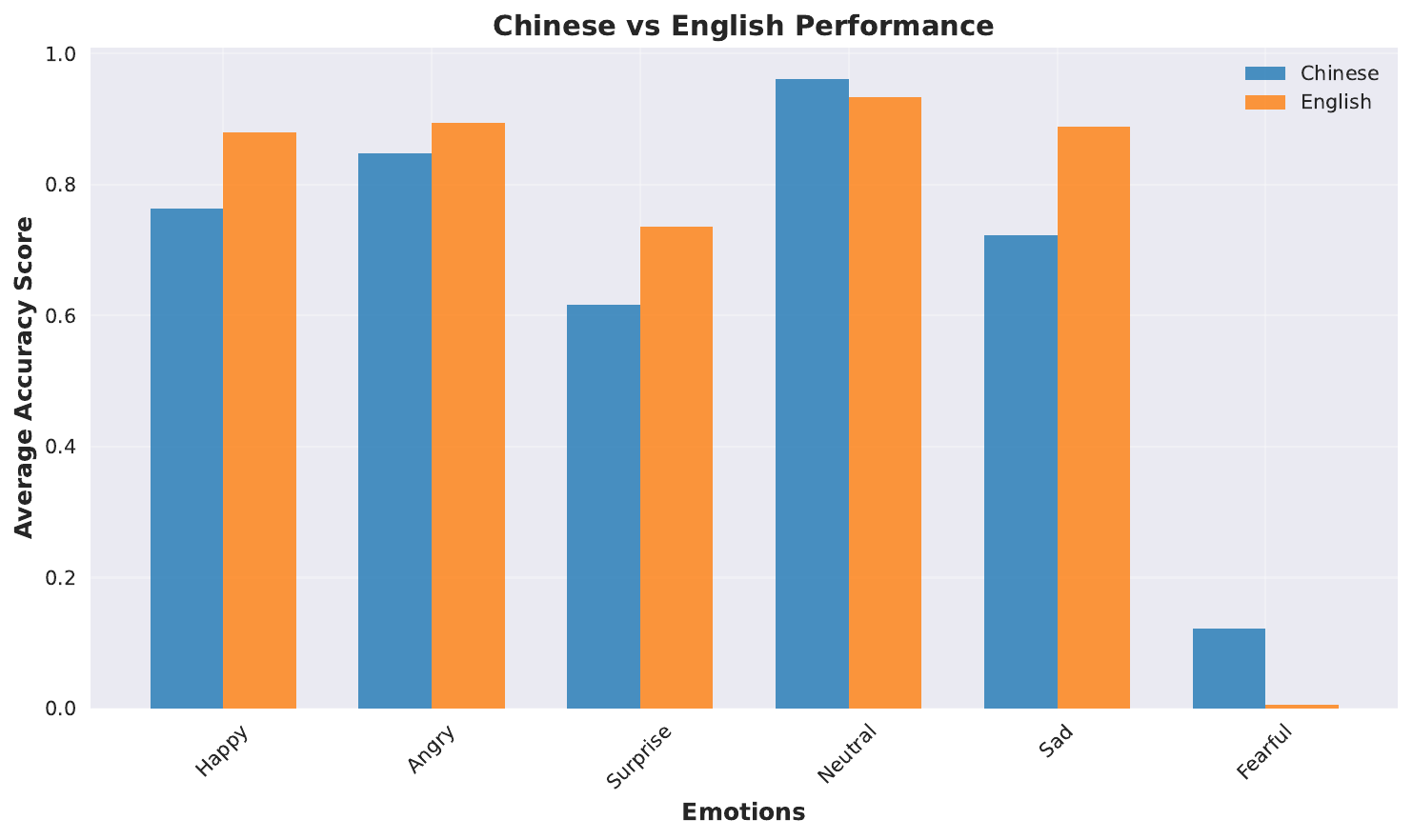}
    \caption{Cross-language performance comparison showing average emotion recognition accuracy between Chinese and English datasets across all model versions.}
    \label{fig:language_analysis_results}
\end{figure}

\paragraph{Language-Specific Patterns}
Figure~\ref{fig:language_analysis_results} shows that Chinese datasets favor happy and angry emotion recognition, while English datasets perform better for neutral and sad emotions. The convergence of performance in advanced model versions (Marco-Voice-v4 and v5) suggests that architectural improvements can reduce language-specific biases and support more universal emotion recognition systems.

\begin{figure}[t]
    \centering
    \includegraphics[width=0.8\linewidth]{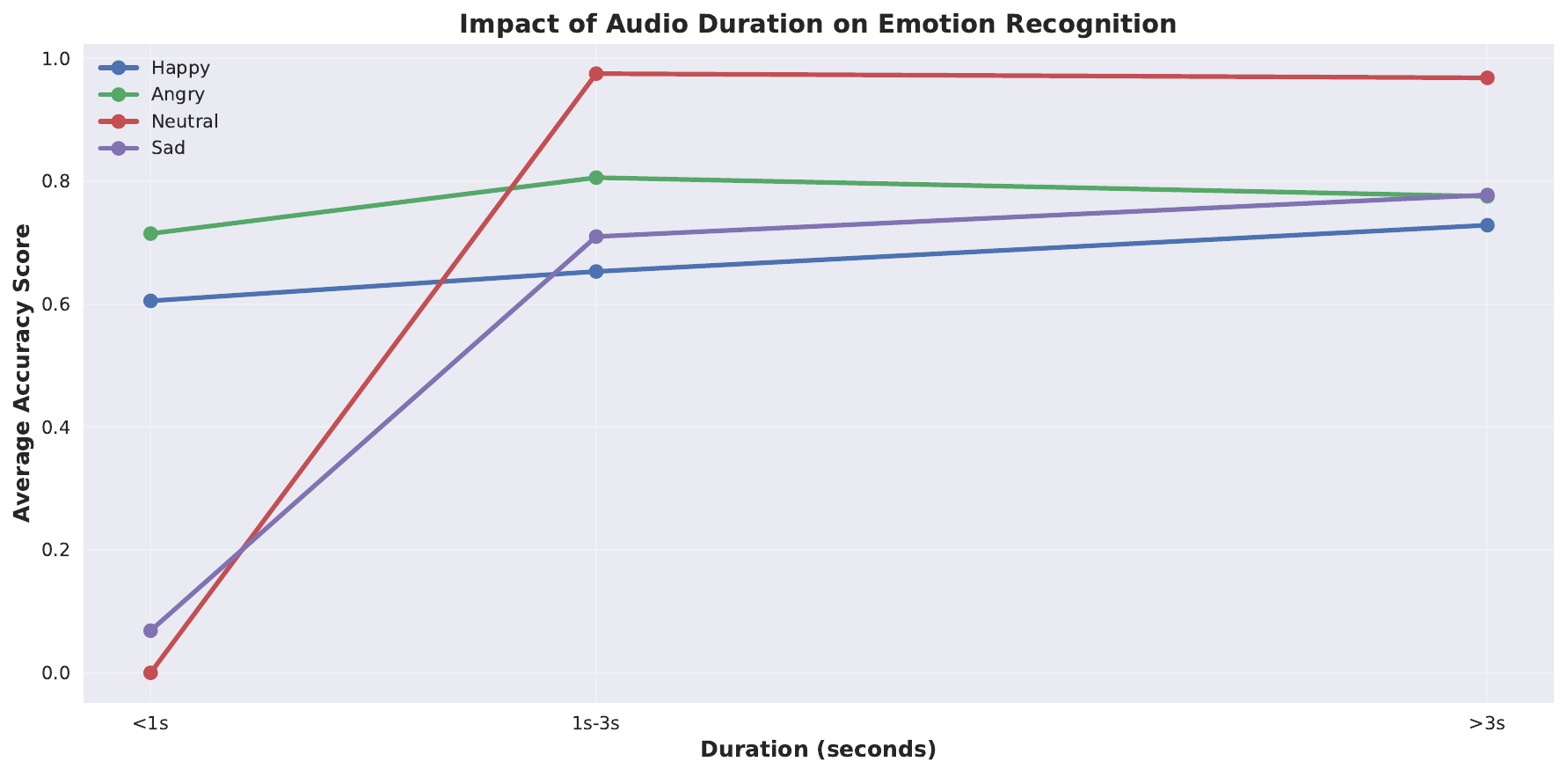}
    \caption{Effect of audio duration on emotion recognition accuracy. Performance is evaluated across three duration categories: short (<1s), medium (1s-3s), and long (>3s) audio segments for four primary emotions.}
    \label{fig:duration_analysis_results}
\end{figure}

\begin{figure}[t]
    \centering
    \includegraphics[width=0.6\linewidth]{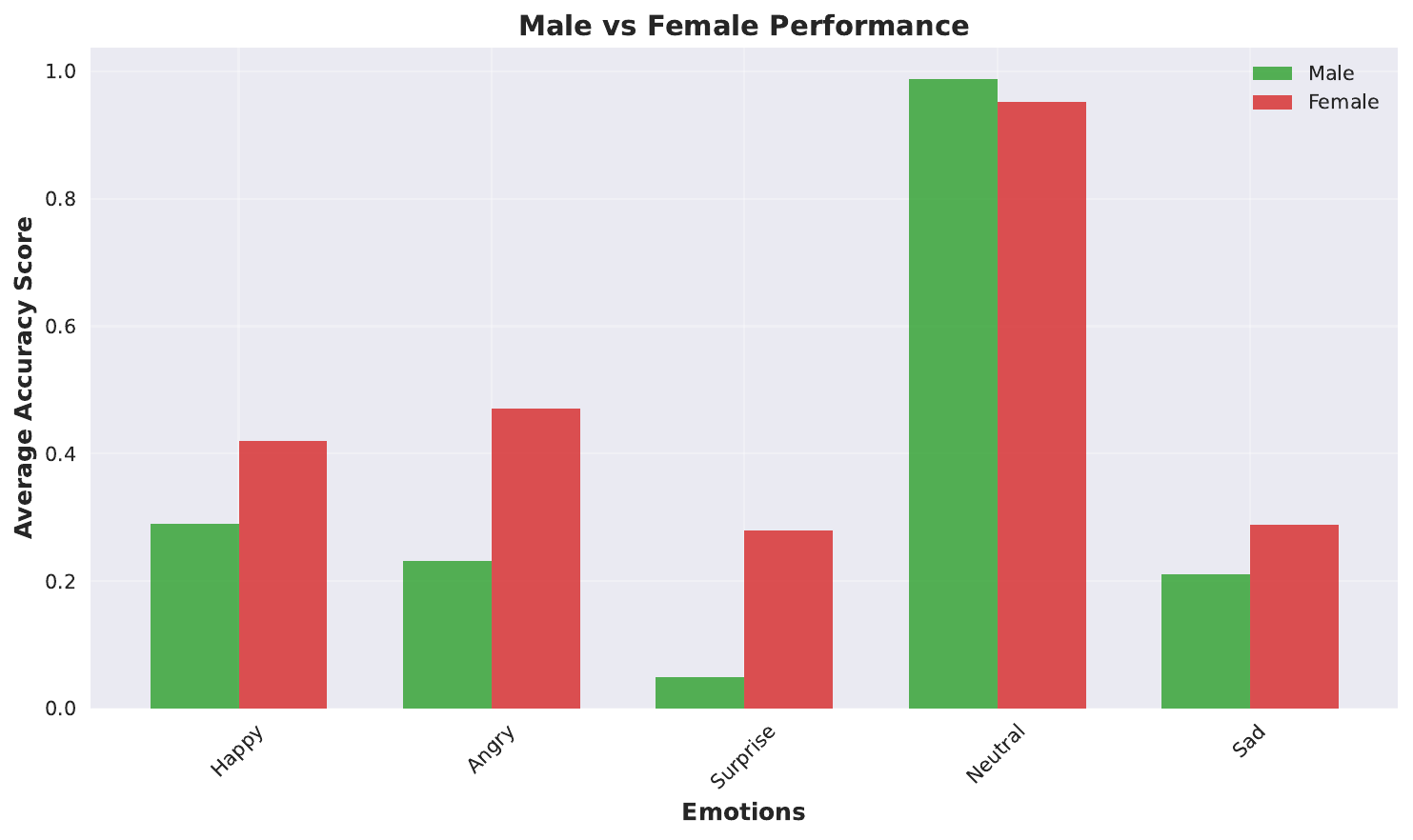}
    \caption{Gender-based performance analysis showing emotion recognition accuracy differences between male and female speakers on the Chinese dataset.}
    \label{fig:gender_analysis_results}
\end{figure}

\paragraph{Duration Impact on Recognition}
Figure~\ref{fig:duration_analysis_results} demonstrates that recognition accuracy increases substantially with audio duration. Short segments (<1s) show poor performance (<0.6), while medium duration (1s-3s) provides optimal efficiency with 0.6-0.8 accuracy. Long segments (>3s) achieve the highest performance but with diminishing returns, indicating 1s-3s as the practical sweet spot for real-time applications.

\paragraph{Gender Performance Disparity}
Figure~\ref{fig:gender_analysis_results} reveals significant gender bias, with male speakers showing substantially lower recognition accuracy across all emotions. Female speakers achieve 0.4+ accuracy for most emotions, while male speakers often fall below 0.2, particularly for surprise and sad emotions. This systematic bias indicates training data imbalances and highlights the need for gender-aware model development.

\section{Discussion}
\subsection{Benefits of Unified Modeling}
Our results demonstrate the advantages of addressing voice cloning and emotional expression within a unified model rather than as separate components. The integrated approach allows the model to learn the subtle interactions between speaker characteristics and emotional expressions, resulting in more natural and consistent speech synthesis.

\subsection{Limitations and Future Work}
Despite the promising results, several limitations remain. First, the current model requires paired emotional speech data, which is scarce for many languages and domains. Future work could explore semi-supervised or self-supervised approaches to reduce this dependency. Second, computational efficiency remains a challenge, particularly for real-time applications. Exploring model compression techniques and optimized inference strategies would make the system more practical for deployment on resource-constrained devices.

\section{Conclusion}
In this paper, we presented Marco-Voice, a multifunctional speech synthesis system that achieves strong performance in voice cloning and emotion controllable speech generation. Through techniques including Rotational Emotion Embedding Integration and Speaker-Emotion Disentanglement as well as other training strategies, our system demonstrates substantial improvements over existing approaches with particular strengths in speaker similarity and emotional expressiveness. The system's unified approach to modeling various speech factors enables more natural and controllable speech synthesis than previous methods that treat these factors in isolation. This work represents an important step toward more expressive and personalized speech synthesis, with potential applications in virtual assistants, accessibility technologies, content creation, and human-computer interaction. Future research directions include expanding language support, reducing data requirements, and optimizing for real-time applications.

\bibliography{references,custom}



\end{document}